# An Evolutional Neural Network framework for Classification of Microarray Data

Maryam Eshraghi Evari, Md Nasir Sulaiman, Amir Rajabi Behjat

**ABSTRACT**

DNA microarray gene-expression data has been widely used to identify cancerous gene signatures. Microarray can increase the accuracy of cancer diagnosis and prognosis. However, analyzing the large amount of gene expression data from microarray chips pose a challenge for current machine learning researches. One of the challenges lie within classification of healthy and cancerous tissues is high dimensionality of gene expressions. High dimensionality decreases the accuracy of the classification. This research aims to apply a hybrid model of Genetic Algorithm and Neural Network to overcome the problem during subset selection of informative genes. Whereby, a Genetic Algorithm (GA) reduced dimensionality during feature selection and then a Multi-Layer perceptron Neural Network (MLP) is applied to classify selected genes. The performance evaluated by considering to the accuracy and the number of selected genes. Experimental results show the proposed method suggested high accuracy and minimum number of selected genes in comparison with other machine learning algorithms.

**KEYWORDS**

DNA Microarray, Dimensionality Reduction, Multi Layer Perceptron, Genetic Algorithm, Classification, Oncology

## I. INTRODUCTION

Microarray is a high throughput technology, provides a format for the measurement of the gene expression levels of thousands of genes simultaneously. It allows expansion of the sample's information to generate detailed expressions of the data for gene regulation and identification [1]. It has been used in studies related to cancer classification, identification of relevant genes for diagnosis or therapy, and investigation of drug effects on cancer prognosis [2].

One exciting result of microarray technology has been the demonstration that patterns of gene expression can distinguish between tumors of different anatomical origins [2], that makes it as an indispensable research tool for understanding the underlying genetic causes of many human diseases [4] and to differentiate between normal and cancerous tissues [8]. To classify multiclass cancer subtype [7], and even to identify new cancer subtypes [5]. Where, the features are genes or, more precisely, their expression levels. In such tasks, supervised classification is quite different from a typical data mining problem, in which every class has a large number of examples. The main task is to propose a classifier of the highest possible quality of classification [3]. However, as Microarray data is characterized by high number of genes (always thousands) the number of samples (always less than 100) is not proportion [6].

The most current Microarray data classification methods are struggling to deal with the curse of dimensionality and accuracy problems.

Machine learning approach includes unified model filtering and ensemble filtering can be used to classify microarray data as if the tissue is cancerous or not []. Previous works on machine learning techniques have benefited from Artificial Neural Network (ANN) [8,9], K-nearest neighbor KNN [10], Naïve Bayesian [..], and Support Vector Machine (SVM) [..] and many other algorithms [] . Nevertheless, only a small number of previous studies focus on parameter optimization for cancer detection but the works exhibit lack of details [19, 20]. Similarly, feature selection experiments in [21, 22, 23] are also lacking in justifications on how the experiment decreases the number of useable features.

In this paper, we propose a Genetic Algorithm-Neural Network hybridization model that classifies gene expression profiles into normal or cancerous tissues. Experimental results on a microarray (Colon cancer microarrays) show that the proposed method presents efficient and effective approach to classify the gene expressions and is superior to other classifiers. We have tried to overcome dimensionality problem during subset selection of informative genes and then classify gene expression Microarray cancer data effectively and efficiently.

The organization of this paper is as follows: Section II presents a study of related work. Section III introduces our proposed technique. In section IV, we describe the results of applying the proposed technique. The paper will be concluded in section V.

## II.    RELATED WORK

Many machine learning methods have been introduced into microarray classification to attempt to learn the gene expression data pattern that can distinguish between different classes of samples [12]. In fact, Machine learning suggests feature selection as a process of feature reduction and variable subset selection for building robust learning models [11]. A generic approach to classifying two types of acute leukemias was introduced in Golub et al. []. Two other systems used for classifying the same microarray dataset was by blending of Support Vector Machine as a classifier, once with Locality Preserving Projection technique (LPP) and the other with Fscore ranking feature selection technique. Both systems result in effectual and powerful classification of gene expression data [12, 13]. Unler [25] focus on Particle Swarm optimization (PSO) algorithm that use subset features encoded in binary strings. However, the resulting low accuracy decreases the efficiency of this method. In [26], the work is to concurrently optimize parameters and feature subset in GA and SVM classifiers without affecting the accuracy rate of SVM classifier. The proposed method performs feature selection and parameter setting in an evolutionary way. Eventually, the resulting accuracy for the GA-based approach with RBF (Radial Basic Function) kernel classifier and the Grid algorithm show a reasonable result with fewer features. Following [26], [27] mixes the behaviors of SVM with GA but at the same time manage to increase the accuracy in comparison with previous study. Ref. [28] examines the use of local search optimization techniques such as Hill Climbing (HC), Simulated Annealing (SA), and Threshold Accepting (TA) as feature selection algorithms and compares the performance of these techniques with Linear Discriminate Analysis (LDA). This study builds K-Nearest Neighbors (K-NN) classifier and the best performance is with SA that reaches to 95.5% accuracy. Artificial Immune System (AIS) has also been explored for feature selection in [24] to generate a set of antibodies from a training set of email messages. The accuracy of 94.6%, however, is slightly lower than in [28]. In the task of classifying gene expressions into cancer and normal tissues, in [5-9] Random Forest (RF) classifier has been applied. The work introduces a new feature selector called topic frequency vectors (TFV) that is able to increase the classification accuracy in comparison with other classifiers such as (SVM, naïve Bayes (NB) and Decision Trees (DT). Findings in [6] conclude that the minimum number of features plays a useful role during classification because it decreases computational time and complexity. In their work, they use two feature selectors that are able to reduce the number of features through Common Vector Approach (CVA) and three-layer Back propagation Neural Network (BPNN) classifier. Ruan [11] forms two-element concentration vector to characterize the according to 'self' concentration and 'non-self' concentration using the 'self' gene library and 'non-self' gene library, respectively [7].

## III. OVERVIEW OF APPROACH

### A. System Description

The proposed hybrid classification method receives high dimensional microarray dataset as input. The system first step is reducing the total number of genes in the input dataset to a smaller subset using Genetic Algorithm as a combined gene selection technique. Then the reduced data will be the data used by the MLP classifiers to assign new samples into their correct classes instead of using the original full data. At this point we can measure and record the test classification accuracy which is equal to the number of correct classified test samples divided by the total number of introduced test samples.

### B. Microarray Gene Expression Datasets

Working with any classification system, any used dataset required to be split into two sub-datasets; a training dataset which the classifier uses to learn and form its learned structure and, a test dataset to see the effectiveness of the proposed system. The proposed classification system work on colon cancer public dataset and can be extended to classify other datasets. Table I contains the details of the dataset. The dataset is available and downloaded from Broad university of Princeton website (www.broadinstitute.org).

| Data Sets | # Genes | # Samples | # Sample per classes | |
|---|---|---|---|---|
| Colon Cancer | 2000 | 62 | Tumor 40 | Normal 22 |

### C. Gene selection

In this work we have used Genetic algorithms (GA), as it known for its capability for optimization. GA works with a set of solutions called population and achieves the optimal solution after a series of repetitive calculations. Then it produces populations of different solutions, which are identified by a chromosome. Each solution acts as the problem, until GA takes suitable results [25]. In addition, GA handles huge search spaces by features exploitation and search exploration, therefore decreases the chance of local optimal solution as compared against other algorithms. Fitness function is one of the important elements in GA that seek to obtain a quality solution from the evaluation step. Crossover and mutation are robust operators that can affect the fitness values randomly [26].

Normally, each feature is scaled between [0, 1] or [-1, 1]. The proposed system of this study scaled features between [-1, 1]. The advantage of scaling is to avoid higher numeric ranges rule the low numeric features. On the other hand, it is caused a high accuracy of classification. In the next step means that feature selection identifies relevant and irrelevant features to make the collection of relational features for training and testing datasets of classifier. In fact this process did training and testing datasets in order to discard unrelated features for avoiding high dimensionality and low classification accuracy. Its main steps are described as below:

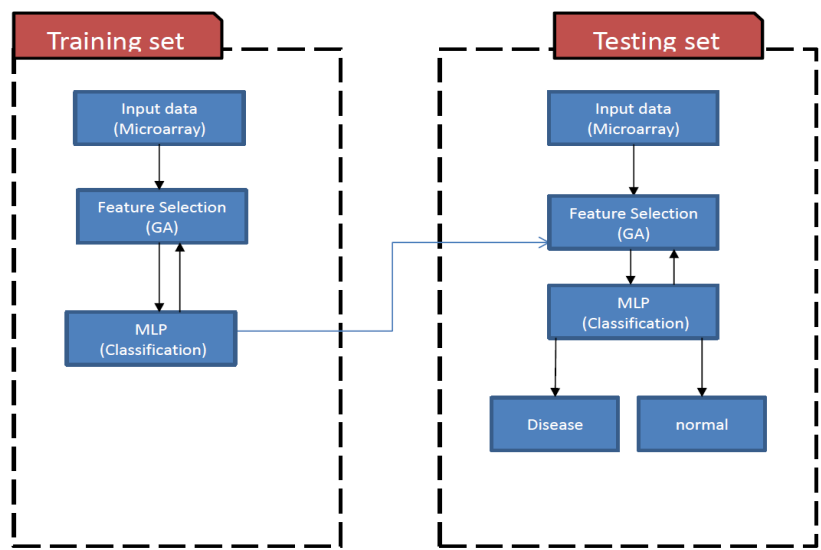

### D. Classification

A Multi-Layer perceptron Neural Network (MLP) is arranged in three layers as a network of nodes, which are the input, hidden, and output layers. Before running the network, a set of weights is randomly arranged for nodes in the network. This will form a new network without knowledge [28]. After preparing the input pattern, the nodes take their attributes from input pattern. In the hidden layer, each attribute is multiplied with a weight by a node. If the value produced is a threshold value, it earns a '1' value, otherwise it takes a value of '0'. This process is followed by output layer, so if the threshold value is passed, a classification process is started by produced input pattern. During the network training the classification process is compared with actual classification and is "backpropagated" within the network, hence the term MLPNN. The nodes of hidden and output layers are produced in this stage to correct their weights responding to each error in classification. The main idea in this network is to classify the best features in linear combinations of the inputs and extracted features from input [24]. The proposed algorithm for training the MLP is shown as follows:

*Generate features classes of cancerous and normal from training samples*

*For each class in training set do*

    *Construct the numerical vector of each class*

*End for*

*Use these numerical vectors for training a MLP.*

*While algorithm is running do*

*If a tissue is received then*

*Arrange the received tissue by numerical vector within the cancerous and normal classes.*

*Predict the label of tissue by training the MLP.*

*End if*

*End while*

### E. Performance Measurement

For evaluation of system performance, we applied the following measurement. Accuracy is defined as the percentage of tissues classified correctly. Accuracy using the binary target datasets can be demonstrated by the false positive (FP) rate, the false negative (FN) rate, the number of tissues that are known as cancerous correctly (TP), and the number of normal tissues (TN). Calculation of accuracy shown in below.

$$Accuracy = Ac = (TP+TN)/(FN+FP+TP+TN)$$

Where TP is the number of Cancerous tissues which are correctly predicted as Cancerous, FN is the number of Cancerous which are predicted as Normal, TN is the number of normal tissues which are predicted as Cancerous and FP is the number of normal tissues which are predicted as Cancerous.

## IV. RESULTS

In this part, different features, which are collected as cancerous and normal features are tested. We tried to find features with best performance to increase the classifier accuracy. Our GA method, which made an optimal binary vector during feature selection, applied in this study decreased the number of useless features and high dimensionality. Each bit is associated with a feature. If a bit of this vector equals 1, then the related feature is allowed to participate during classification, if the bit is a 0, then the feature does not participate. At the end of feature selection, the number of features is decreased from 2000 to 2. Furthermore, for measuring performance of cancerous tissues detection, 20 runs are done to measure the performance exactly. Thus in each iteration, 90% of the data was used for training while the rest 10% are used for testing. The MLP classifies the testing tissues into cancerous or normal tissues classes. Note that the number of nodes equals to the size of input vector. In this study, an input vector is collected from a set of "0" and "1". The nodes of hidden layer are tested from 3 to 15. Output layer followed two nodes, first node indicates cancerous tissues and second node is normal tissues. The network is trained for a maximum of 60 epochs to 0.01 of error goal. Table 1 shows the performances of Linear Discriminant [20-21], SVM (Linear Kernel) [22], SVM (RBF Kernel) [23], BP Neural Network [24] and MLP proposed in this study on Ling-Spam dataset respectively. All these results are obtained by using 10-fold validation and 50 to 100 features. For example, SVM uses information gain (IG) as feature selection criteria, and the best scoring 256 features are chosen.

## V.     ANALYSIS

Two criteria have been used to evaluate the effectiveness of the proposed approach included: number of selected features (samples) and predicative accuracy. Select the smallest number of samples that can achieve the highest predictive accuracy. The best result is to select the smallest number of samples which can achieve the highest predictive accuracy. Following table presents our final results compare to other models (SVM and Naïve basin) results.

| Method | Accuracy (%) | # Feature |
|---|---|---|
| SVM | 93.55% | 2 |
| Naïve Basian | 93.55% | 3 |
| Proposed Method | 99.87% | 2 |

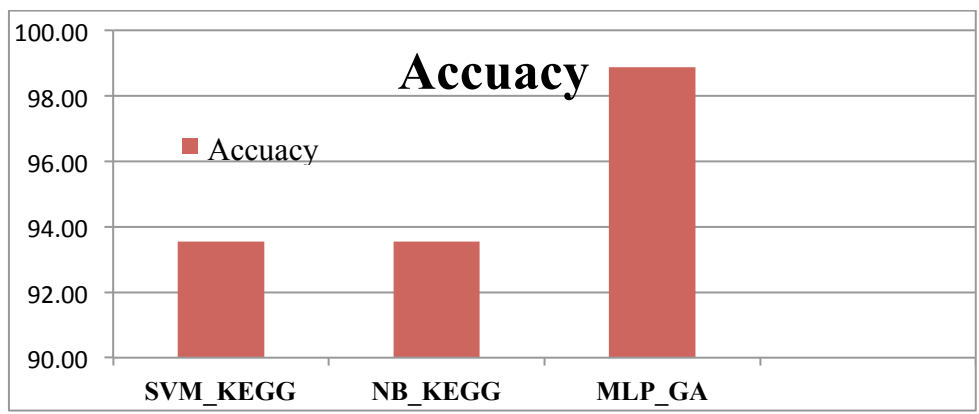

## VI. CONCLUSION

In this paper, an MLP Neural Network is used for Cancerous tissues detection. To increase accuracy of MLP classifier, feature selection employed two methods, which are heuristics and GA to form a vector with useful features as the input vector to the MLP classifier. This in turn decreased the feature dimensionality through removal of useless features. Experiments with the public Colon gene expression dataset has proven that the proposed classifier with the selected feature selection methods does not only achieved high speed but the classification experiment also achieved 99.87% accuracy only by using declined concentration feature vector. This study showed experimental results for features of body and subject based on GA and MLP classifier.